\documentclass{article}

\usepackage{microtype}
\usepackage{graphicx}
\usepackage{subfigure}
\usepackage{booktabs} 

\usepackage{hyperref}

\usepackage[accepted]{icml2020}

\icmltitlerunning{An Open-World Simulated Environment for Developmental Robotics}

\begin{document}

\twocolumn[
\icmltitle{An Open-World Simulated Environment for Developmental Robotics}



\icmlsetsymbol{equal}{*}

\begin{icmlauthorlist}
\icmlauthor{SM Mazharul Islam}{uta}
\icmlauthor{Md Ashaduzzaman Rubel Mondol}{uta}
\icmlauthor{Aishwarya Pothula}{uta}
\icmlauthor{Deokgun Park}{uta}
\end{icmlauthorlist}


\icmlaffiliation{uta}{Department of Computer Science and Engineering, University of Texas, Arlington, Texas USA}

\icmlcorrespondingauthor{Deokgun Park}{deokgun.park@uta.edu}

\icmlkeywords{Developmental Robotics, Sensorimotor Development}

\vskip 0.4in]



\printAffiliationsAndNotice{} 

\begin{abstract}
As the current trend of artificial intelligence is  shifting towards self-supervised learning, conventional norms such as highly curated domain-specific data, application-specific learning models, extrinsic reward based learning policies etc. might not provide with the suitable ground for such developments. In this paper, we introduce SEDRo, a Simulated Environment for Developmental Robotics which allows a learning agent to have similar experiences that a human infant goes through from the fetus stage up to 12 months. A  series  of  simulated  tests  based  on  developmental  psychology will  be  used  to  evaluate  the  progress  of  a  learning  model.
\end{abstract}

\section{Introduction}
\label{submission}

Artificial Intelligence (AI) has accomplished great progresses in the field of application-specific learning. State of the art models in computer vision and natural language processing are performing with near human-level accuracy. However, majority of these breakthroughs are occurring in the context of supervised learning and reinforcement learning where there exists ground truth or reward signals. As a result, more often than not, proposed models are specifically designed to acquire a single problem-solving skill which can not be generalized well on different problem-solving domains. Feeding with refined and focused data sets rather than using diverse and noisy sources further exacerbates this issue.  High dependency on explicit reward mechanisms to guide the learning process makes conventional models over-fitted to a specific task, and it becomes exponentially difficult to generalize such models over multiple tasks. 

\begin{figure}[t]
    \centering
    \resizebox{\columnwidth}{!}{\includegraphics{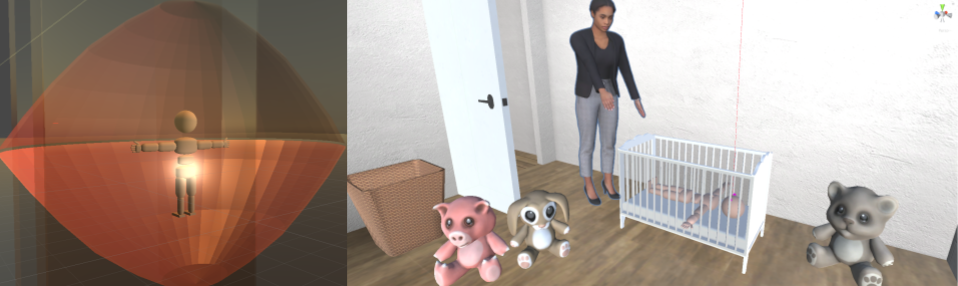}}
    \caption{\small Screenshots of SEDRo environment. The environment simulates a fetus in the womb (left). An infant (learning agent) in the crib with a mother character (right). }
    \label{fig:my_label}
\end{figure}

To address these issues, using a common learning mechanism to conduct multiple tasks can be an effective regularization. Either a virtual or a physical environment is required which would allow a learning agent to interact with the surrounding objects. Through such interactions an agent can gather multi-modal data about its environment which is how a biological system usually explores. 
Due to issues like manufacturing and maintenance cost, reproducibility, experiment time etc., often times a virtually simulated environment is preferred over a physical one. 

In this paper, we present our ongoing work on the Simulated Environment for Developmental Robotics (SEDRo). 
SEDRo  provides diverse experiences similar to human infants ranging from the fetus stage to 12 months age. Also there is a caregiver character which will assist an agent with cognitive bootstrapping. 
The objects (e.g. crib, toys, windows, walls etc.) will respond to the agent's interaction through physics simulations from the Unity 3D game engine. 
The physical abilities of the agent and the efficacy of its sensors are subject to gradual enhancements, which follows the development of a human infant. To evaluate the intellectual progress of the agent at different stages, several developmental-psychology based tests are provided.

\section{Proposed Environment}

\begin{figure}[h!]
  \centering
  \resizebox{0.97\columnwidth}{!}{\includegraphics{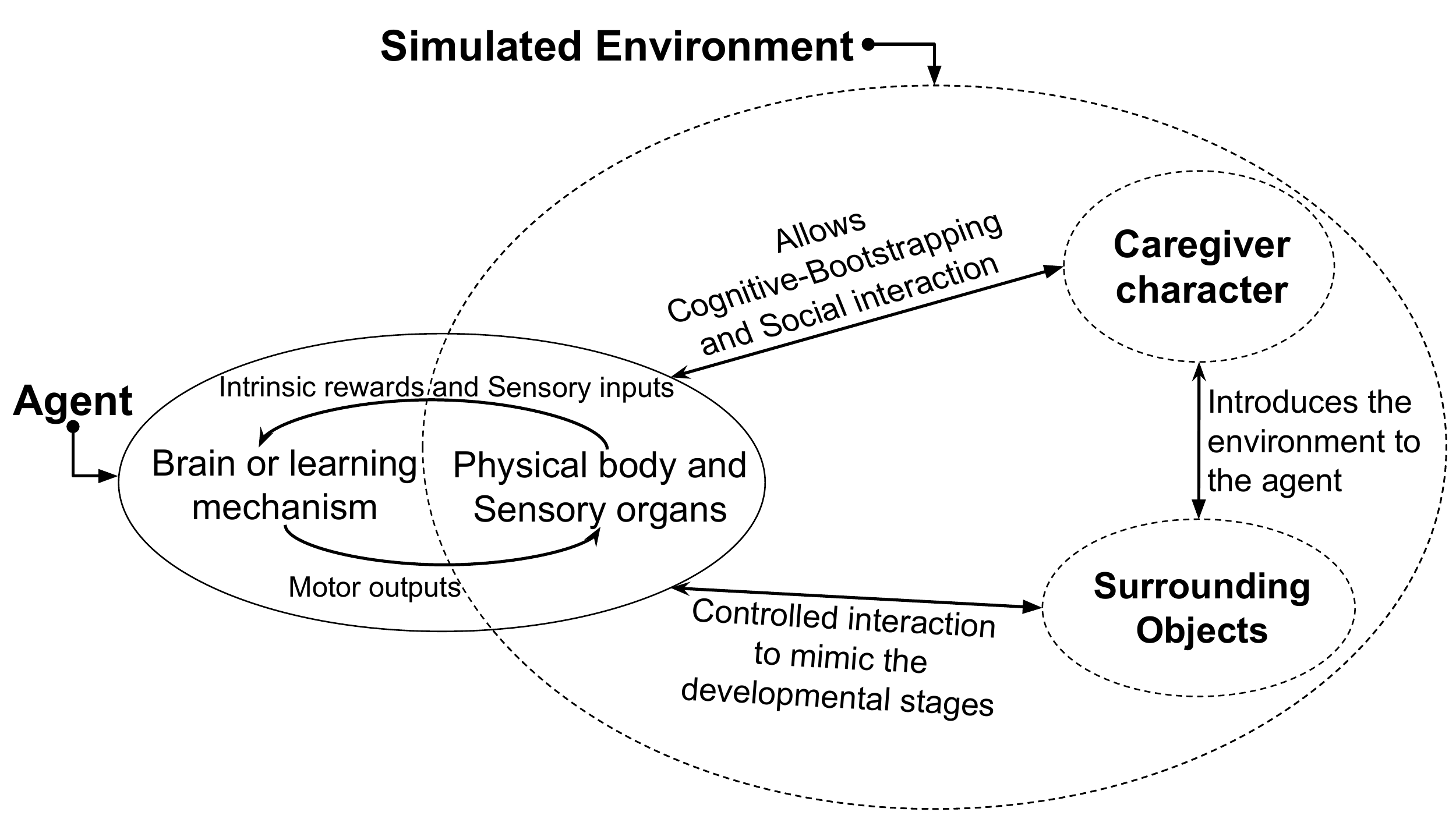}}
  \caption{\small Ecosystem of SEDRo environment.}
  \label{fig:big_picture}
\end{figure}

In this section, we present the proposed Simulated Environment for Developmental Robotics or SEDRo. The primary components of SEDRo and their relations are shown in Fig.~\ref{fig:big_picture}. 
The two main components in SEDRo are the learning agent (with solid border), the simulated environment (with dashed border). 
Interactions between the agent and the caregiver allow cognitive bootstrapping and social-learning, while interactions between the agent and the surrounding objects are increased gradually as the agent gets into more developed stages. 

SEDRo does not provide with any explicit reward mechanisms that might drive an agent to learn a particular skill. Instead, the responsibility of generating rewards belong to the agent itself. 
As an example, the energy level of the agent (i.e. food in its stomach) will be represented by a real number. The agent itself is responsible for generating a positive reward upon receiving food from the environment or a negative reward during starvation. Hence, the body of the agent is also part of the environment and the term "agent" refers to the brain only.

In the current version, an agent is fed with sensory inputs recorded by touch, vision, acceleration, gravity and proprioceptors sensors. An agent can control its eye movements with 3 degree of freedom (DOF) and has access to both central and peripheral vision. 
Touch sensors are spread all over the agent's body with higher density in face and limb regions, and provide a binary feature which represents if a contact has been made. 
In each hand, the motor output vectors constitute muscle torques, which will determine the 53 motors, including the 9 DOF.

\section{Evaluation Framework}

\begin{figure}[t!]
  \centering
  \resizebox{0.7\columnwidth}{!}{\includegraphics{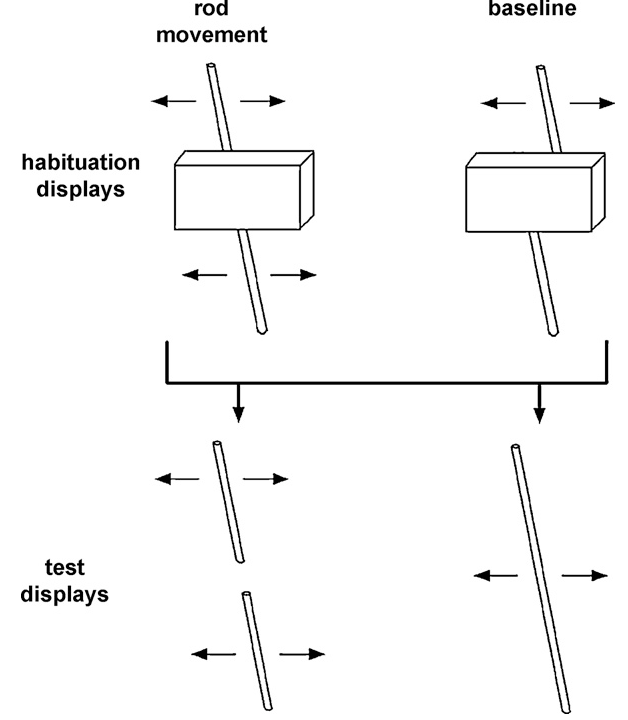}}
  \caption{{\small Rod and Box experiment from ~\cite{kellman1983perception}.} }
  \label{fig:evaluation}
\end{figure}

An artificial agent within SEDRo encounters similar experiences of human infants from fetus stage up to 12 months of age, and infants from this age range are not capable of verbal communication. 
In developmental psychology, to effectively measure the intellectual developments of such non-verbal human infants, a range of experiments have been proposed~\cite{cangelosi2015developmental}.
The scope of these experiments are broad enough to cover a diverse behavioral spectrum and multiple sensory inputs such as vision, joint attention, motor, language, reasoning etc.
Hence, within SEDRo environment, we use simulations of such established experiments as means to evaluate an artificial agent's intellectual capabilities in different domains. 

Fig.~\ref{fig:evaluation} illustrates one of such experiments, known as rod-and-box experiment ~\cite{kellman1983perception}. The beginning of this experiment is the habituation phase where a rod is laterally moved and it's central portion is hidden behind a large, solid screen or box. As infants get habituated to this scene, which can be predicted by their below-average eye-gaze time, one of the two following scenes are shown. During this post-habituation stage, in one case, a solid rod is shown without the solid box. In another case, two segments of rod are shown as if the initial rod has been been broken at the center. The reactions from the infants can be compared based on their average eye-gaze time. Since new-born infants do not have perceptual completion ~\cite{slater1996newborn}, they tend to spend more time looking at the solid rod as that is more familiar to the original rod behind a box scene. However, after four months infants start to develop the perceptual completion skill and spend longer time looking into the broken rod scene. Within SEDRo, we will compare the average eye-gaze time from artificial agents and response from human infants to measure development of perceptual skills.

\bibliography{references.bib}

\begin{thebibliography}{3}
\providecommand{\natexlab}[1]{#1}
\providecommand{\url}[1]{\texttt{#1}}
\expandafter\ifx\csname urlstyle\endcsname\relax
  \providecommand{\doi}[1]{doi: #1}\else
  \providecommand{\doi}{doi: \begingroup \urlstyle{rm}\Url}\fi

\bibitem[Cangelosi \& Schlesinger(2015)Cangelosi and
  Schlesinger]{cangelosi2015developmental}
Cangelosi, A. and Schlesinger, M.
\newblock \emph{Developmental robotics: From babies to robots}.
\newblock MIT press, 2015.

\bibitem[Kellman \& Spelke(1983)Kellman and Spelke]{kellman1983perception}
Kellman, P.~J. and Spelke, E.~S.
\newblock Perception of partly occluded objects in infancy.
\newblock \emph{Cognitive psychology}, 15\penalty0 (4):\penalty0 483--524,
  1983.

\bibitem[Slater et~al.(1996)Slater, Johnson, Brown, and
  Badenoch]{slater1996newborn}
Slater, A., Johnson, S.~P., Brown, E., and Badenoch, M.
\newblock Newborn infant's perception of partly occluded objects.
\newblock \emph{Infant Behavior and Development}, 19\penalty0 (1):\penalty0
  145--148, 1996.

\end{thebibliography}
\bibliographystyle{icml2020}


\end{document}